\documentclass{article}

\usepackage{arxiv}

\usepackage[utf8]{inputenc} % allow utf-8 input
\usepackage[T1]{fontenc}    % use 8-bit T1 fonts
\usepackage{hyperref}       % hyperlinks
\usepackage{url}            % simple URL typesetting
\usepackage{booktabs}       % professional-quality tables
\usepackage{amsfonts}       % blackboard math symbols
\usepackage{nicefrac}       % compact symbols for 1/2, etc.
\usepackage{microtype}      % microtypography
\usepackage{lipsum}
\usepackage{graphicx}
\graphicspath{./Figure/}
\usepackage{subcaption}
\usepackage{ragged2e}
\usepackage{tabularx}
\usepackage{amsmath}
\usepackage{mathtools}
\usepackage{cuted}
\usepackage{amssymb}
\usepackage{lipsum}
\usepackage{nccmath}
\usepackage{autobreak}
\usepackage{makecell}
\usepackage{xurl}
\usepackage{xcolor}

\title{Modeling subjectivity (by Mimicking Annotator Annotation) in toxic comment identification across diverse communities \\
{\footnotesize \textcolor{red}{WARNING: This paper contains comments that are offensive by nature.}}
}
\author{
 Senjuti Dutta \\
  University of Tennessee, Knoxville\\
  Knoxville, TN  \\
  \texttt{sdutta6@vols.utk.edu} \\
   \And
 Sid Mittal \\
  Google LLC \\
 Mountain View, CA \\
  \texttt{sidmittal@google.com} \\
  \And
 Sherol Chen \\
  Google LLC \\
  Mountain View, CA\\
  \texttt{sherol@google.com} \\
 \And
Deepak Ramachandran \\
    Google LLC \\
    Mountain View, CA\\
    \texttt{ramachandrand@google.com} \\
    \And
Ravi Rajakumar \\
    Google LLC \\
    Mountain View, CA\\
     \texttt{ravirajakumar@google.com} \\
    \And
Ian Kivlichan \\
    Google LLC \\
    Mountain View, CA\\
     \texttt{Ian.Kivlichan@gmail.com} \\
    \And
Sunny Mak \\
    Google LLC  \\
    Mountain View, CA\\
     \texttt{smak@google.com} \\
    \And 
Alena Butryna \\
    Google LLC \\
    Mountain View, CA\\
     \texttt{alenab@google.com} \\
    \And
Praveen Paritosh \\
    Google LLC  \\
    Mountain View, CA\\
     \texttt{pkparitosh@gmail.com} \\
}
\begin{document}
\maketitle
\begin{abstract}
The prevalence and impact of toxic discussions online have made content moderation crucial. Automated systems can play a vital role in identifying toxicity, and reducing the reliance on human moderation. Nevertheless, identifying toxic comments for diverse communities continues to present challenges that are addressed in this paper. The two-part goal of this study is to (1) identify intuitive variances from annotator disagreement using quantitative analysis and (2) model the subjectivity of these viewpoints. To achieve our goal, we published a new dataset\footnote{\url{https://github.com/XXX}} with expert annotators' annotations and used two other public datasets to identify the subjectivity of toxicity. Then leveraging the Large Language Model (LLM), we evaluate the model’s ability to mimic diverse viewpoints on toxicity by varying the size of the training data and utilizing the same set of annotators as the test set used during model training and a separate set of annotators as the test set.  
We conclude that subjectivity is evident across all annotator groups, demonstrating the shortcomings of majority-rule voting. Moving forward, subjective annotations should serve as ground truth labels for training models for domains like toxicity in diverse communities.  

\end{abstract}

\section{Introduction}

Online platforms play a vital role in communication, creation, discussion, and collaboration. They allow users to discuss various topics by building communities with shared interests \cite{mohan2017impact}. However, communities can also bring out negativity through their community members engaging in toxic behavior by spreading hate speech, using offensive language, fake news, expressing personal attacks, cyberbullying about a religion, person, group, or public figure, etc.\cite{wulczyn2017ex, zhao2016automatic}.
% As online social media and communication platforms recognized the need to combat toxicity, they implemented a set of rules and norms that users must follow. 
% Human moderators are still used in a system  that monitors and removes toxic posts by using regular expressions and blacklists to catch inappropriate language. Effectively moderating toxic postings that target users with various forms of harassment and rudeness become increasingly complex due to the rise in user involvement in online communities and the harsh remarks left while hiding behind anonymity. 
The prevalence and impact of toxic online discussions have led both industrial and research communities to combat toxicity beyond human moderation, through automated systems and sophisticated ML filtering. 

% to alleviate the need for human moderation.

%especially by using automated systems for identifying toxicity and alleviating the need for human moderation.
% It has become crucial to moderate online content due to the prevalence and impact of toxic discussion, especially by using automated systems for identifying toxicity, and alleviating the need for human moderation.
Human-annotated labels are the primary empirical source of ground truth for training toxicity detection models. 
However, a key limiting factor in using datasets with toxicity annotations is the low consistency and reliability of the labels due to their intrinsic subjectivity \cite{goyal2022your,arhin2021ground}. 
 % However, one of the main challenges with subjective datasets is the consistency and reliability of labels provided by annotators \cite{goyal2022your,arhin2021ground}. 
 % Toxicity is a subjective property of sentences. 
To gain a better grasp of the subjective nature of toxicity, let us examine an illustrative example:
 \begin{quote}
 \emph{``Another great `zinger'. You're an amusing fellow. I bet you sway all kinds of folks to your side of an issue.''} 
 \end{quote}
 If two different annotators are asked to annotate the toxicity of the above comment, they might express completely different but valid opinions.
 This comment could be interpreted as toxic to annotators assuming that it is harassment to the interlocutor or it could be not toxic to annotators assuming it is sardonic but not rude or disrespectful. 
This can have a negative impact on the fairness and accuracy of the models trained to classify toxicity due to its subjective nature. 
Thus, in this paper, we aim to understand the subjectivity of toxic datasets and explore the usage and limitations of toxicity annotations in model training. Consequently, this approach will aid in diminishing toxic harassment and restraining malicious posts by identifying and addressing toxic comments across diverse communities. 
 While the subjectivity of toxicity is acknowledged, the extent of subjectivity remains unclear when toxic comments are annotated by both expert and non-expert annotators, as well as across different annotator groups and timelines.
Therefore in this paper, we investigate subjectivity through quantitative analysis.
There have been many publicly available toxic datasets ~\cite{davidson2017automated, mubarak2017abusive, wiegand2018overview,zampieri2019predicting,goyal2022your,leonardelli2021agreeing}. However, these datasets are not annotated by annotators who have experience in annotating subjective datasets. 
Moreover, these datasets lack a crucial element, the inclusion of annotators' rationales behind their judgments.
% as well as these datasets  do not take into account whether a toxic comment would cause an annotator to leave the discussion.

We fill the gap first by publishing a dataset by recruiting annotators who possess expertise in subjective datasets, whom we refer to as ``expert annotators.'' 
We explicitly incorporate the annotators' rationales during the dataset creation process, thereby addressing the limitation above.
We used two publicly available datasets\footnote{\url{https://www.kaggle.com/c/jigsaw-unintended-bias-in-toxicity-classification}}, \footnote{\url{https://www.kaggle.com/datasets/google/jigsaw-specialized-rater-pools-dataset}} along with our created dataset \footnote{\url{https://github.com/XXX}} for quantitative analysis in the context of toxicity.
%\footnote{https://github.com/google-research-datasets/toxicity-replication} 
% where comments are annotated across three timelines. 
% Therefore next we investigate subjectivity through quantitative analysis in the context of toxicity for three different datasets\footnote{https://www.kaggle.com/c/jigsaw-unintended-bias-in-toxicity-classification, https://www.kaggle.com/datasets/google/jigsaw-specialized-rater-pools-dataset, https://github.com/google-research-datasets/toxicity-replication/blob/main/toxicity\textunderscore rep \textunderscore dataset.csv} where comments are annotated across three timelines. 
% Section \ref{sec: understanding_subjectivity} presents the methodology employed and discusses the results derived from this quantitative analysis.
Then we model the subjectivity of diverse viewpoints using an LLM framework. 
% Through this, we assess the model's ability in understanding different annotators' views on toxicity.
% For this,  we present an illustration within the framework of a large language model (LLM) to assess the model's ability to capture individual annotators' perspectives on toxicity.
% There we aim to assess the model's ability to capture individual annotators' perspectives on toxicity. 
To assess the model's ability to capture individual annotators' perspectives on toxicity, we employ two methods: (1) varying the size of the training data and (2) utilizing both the same set of annotators as the test set used during model training and a separate set of annotators as the test set. 
% In the first approach, we manipulate the size of the training data to examine how it influences LLM's ability to learn and comprehend the unique perspectives of annotators regarding toxicity. By systematically adjusting the training data size, we can analyze the model's sensitivity and its capacity to effectively capture individual annotator viewpoints.
% Furthermore, we utilize two sets of annotators for the test set. The first set consists of annotators used during model training, allowing us to evaluate the model's performance in reproducing the perspectives it was exposed to during training. The second set comprises a completely new group of annotators, enabling us to assess the model's generalization abilities by evaluating its understanding and adaptation to previously unseen perspectives on toxicity.
% This approach enables us to comprehensively evaluate and analyze the performance of the model with respect to capturing subjectivity across different training data sizes and diverse annotator perspectives.
These methods allow us to thoroughly assess how well the model performs in understanding various viewpoints, especially when trained on different amounts of data and with input from a variety of annotators.

In the remainder of the paper, we first provide an overview of related work in Section~\ref{sec:related_work}.
Section~\ref{sec:understanding_subjectivity} presents the methodology employed and discusses the results derived from this quantitative analysis.
In Section~\ref{sec:personalizing_subjectivity_llm}, we describe how we model subjectivity and detail the experiments conducted using an LLM framework.
% Section~\ref{sec:personalizing_subjectivity_llm} presents the modeling of subjectivity by demonstrating the conducted experiments for mimicking the subjectivity of annotators using an LLM framework. 
Next, we conclude with takeaways from the paper in Section~\ref{sec:conclusion}. Finally Section~\ref{sec:ethical} addresses the limitations and ethical considerations of our work, as well as broader implications. 
% Section \ref{sec:conclusion} concludes with takeaways from the paper. Section \ref{sec:ethical} addresses the limitations of our work, and ethical considerations of our work, as well as discusses the broader impacts of our work's implications. 
\section{Related work}
\label{sec:related_work}
To position our research we present a brief summary of the related work in two areas: metrics for subjective datasets, and previous work reflecting disagreement of toxicity annotation.

\subsection{Metrics for handling annotator agreement }
Previous studies have presented various metrics for measuring agreement among annotators. In the context of crowdsourcing-based approaches, inter-rater reliability (IRR) is commonly utilized as a measure to assess the level of agreement when collecting annotated data. 
Inter-rater reliability (IRR) is commonly utilized as a measure to assess the level of agreement when collecting annotated data. The IRR metrics encompass a wide range of coefficients, accommodating different experimental scenarios such as varying numbers of raters, rating scales,  agreement definitions, different types of crowdsourcing tasks, and assumptions regarding rater interchangeability. Examples of these metrics include Scott's pi \cite{scott1955reliability}, Cohen's kappa \cite{cohen1960coefficient}, Siegel and Castellan’s kappa \cite{siegel1957nonparametric}, Fleiss's kappa \cite{fleiss1971measuring}, Byrt et al.'s kappa \cite{byrt1993bias}, Krippendorff's alpha \cite{krippendorff2011computing}, CrowdTruth metrics \cite{dumitrache2018crowdtruth}, Jury Learning \cite{gordon2022jury}, etc.
% is a versatile reliability coefficient that can handle various scenarios, including different levels of measurement and missing data.
% Crowdtruth Metrics \cite{dumitrache2018crowdtruth}, etc. 
% Additionally, there are alternative metrics like CrowdTruth metrics \cite{dumitrache2018crowdtruth} that can be employed in crowdsourcing annotation tasks to capture and interpret inter-annotator disagreement. These metrics are particularly useful in crowdsourcing tasks involving semantic annotation, where data ambiguity and multiple perspectives on information examples are prevalent.
% Inter-rater reliability is also assessed using rank correlation coefficients such as Spearman's rho or Kendall's tau \cite{kendall1948rank} or Kendall's coefficient of concordance \cite{kendall1970rank}.
% There are many other ways also to measure 
Additionally, to measure the agreement between annotators across different groups,  \cite{wong2021cross} propose cross-replication reliability (xRR) and normalized xRR. 
\subsection{Disagreement of toxicity annotation}

% \subsection{}

Prior studies have examined the evaluation of toxic datasets, revealing notable instances of disagreement among annotators. For instance, in the context of labeling comments as ``toxic'' for Wikipedia talk page comments, annotators exhibited Krippendorff's alpha 0.5, indicating moderate disagreement among them  \cite{binns2017like}.
Moreover, \cite{ross2017measuring} found that researchers who were already familiar with the definition of hate speech exhibited relatively low agreement (Krippendorff's alpha of 0.38).
% comment moderation on the /r/science subreddit, moderators exhibited a Fleiss' kappa score of 0.46 when manually annotating comments, indicating moderate disagreement among them \cite{lucas2019understanding}. Another investigation i.
When annotating the Civil Comments dataset, African American and LGBTQ groups considered comments to be more toxic than a control rater pool composed of individuals who are neither African American nor LGBTQ \cite{goyal2022your}.
Similarly, a study focusing on labeling tweets on a four-point toxicity scale from a corpus of 20,000 English-language tweets revealed relatively low inter-rater agreement (Fleiss' kappa of 0.25), even among raters who had previously experienced online harassment \cite{diaz2022crowdworksheets}. Collectively, these previous works demonstrate significant disagreement among annotators when undertaking toxicity tasks.
% Overall this prior work demonstrates that there is significant disagreement between annotators in toxicity annotations. 
% Previous research has shown assessments of subjective datasets. For example, Moderators of a popular Reddit subreddit often disagree with each other, showing a Fleiss' kappa for comments annotated by two moderators of 0.46 for manually moderating comments on the /r/science subreddit \cite{lucas2019understanding}. Inter-rater agreement is significantly lower  for women than for men (both under Krippendorff's alpha 0.5) regarding labeling comments as `toxic' for Wikipedia talk page comments \cite{binns2017like}.  African American groups and LGBTQ groups consider a comment as more toxic than the control rater pool (who are neither African American nor LGBTQ) in annotating the Civil Comments dataset \cite{goyal2022your}.
% Similarly, in a study of labeling each tweet in the four-point toxicity scale from the Twitter corpus of 20,000 English-language tweets, the authors found relatively low inter-rater agreement (Fleiss' kappa is 0.25) even among raters who had previously experienced online harassment \cite{diaz2022crowdworksheets}.
% Overall this prior work demonstrates that there is significant disagreement between annotators in toxicity annotations. 
However, it remains unclear whether similar types and amounts of subjectivity are prevalent for the same datasets across different timelines and also among expert annotators who are specifically trained for the task of annotating subjective datasets.

\section{Understanding subjectivity of toxicity across three dataset versions}

\label{sec:understanding_subjectivity}
%\subsection{Method }

 % In this section, we first describe two publicly available datasets that cover different toxic categories annotated by non-experts. Next, 
 In this section, we first describe two publicly available datasets and then discuss our published dataset as one of the main contributions of this paper, which includes annotations from expert annotators. Subsequently, we present our task design and the analysis of the collected data. Ultimately, we delve into a discussion of the results.
 % Thereafter, we describe our task design and the analysis of collected data.
 % Finally, we discuss the result which helps to understand the subjectivity of toxicity using quantitative analysis.

\subsubsection{Datasets}
% There have been many publicly available toxic datasets \cite{davidson2017automated, mubarak2017abusive, wiegand2018overview,zampieri2019predicting,goyal2022your,leonardelli2021agreeing}. However, these datasets lack a crucial element: the inclusion of annotators' rationales behind their judgments as well as these datasets  do not take into account whether a toxic comment would cause an annotator to leave the discussion.
% % % Moreover, these datasets often lack the involvement of annotators with experience in working with subjective datasets. Moreover, the existing publicly available datasets do not take into account whether a toxic comment would cause an annotator to leave the discussion.
% In contrast, our paper aims to fill this gap by recruiting annotators who possess expertise in subjective datasets, whom we refer to as ``experts.'' We explicitly incorporate the annotators' rationales during the dataset creation process, thereby addressing the aforementioned limitation.  Furthermore, to assess the annotators' inclination to leave the discussion, we employed a binary choice format, inquiring whether the comment would prompt them to leave. By leveraging the insights of these experts, our approach enhances the transparency and interpretability of the toxic dataset, leading to more meaningful and reliable annotations.
Here we explain all three datasets that are used in the paper. The first two datasets are publicly available: Civil Comment Toxicity Kaggle (CCTK) Dataset\footnote{\url{https://www.kaggle.com/c/jigsaw-unintended-bias-in-toxicity-classification}}, we refer as \textit{2017 dataset} and Rater Pool Dataset\footnote{\url{https://www.kaggle.com/datasets/google/jigsaw-specialized-rater-pools-dataset}}, addressed as \textit{2022 dataset}. Both the 2017 and 2022 datasets are described in Appendix~\ref{sec:appendix_dataset}). The third dataset is the one that we created and published\footnote{\url{https://github.com/XXX}}. We address the dataset as the \textit{2023 dataset (expert annotators)}. These three datasets are all civil comments annotated by different annotators across three timelines.

\emph{2023 dataset(expert annotators)}: 
We selected a total of 50 comments from the 2017 and 2022 datasets, where 30 random comments were bucketed by the amount of disagreement between annotators from the 2017 dataset, and 20 random comments were selected from the 2022 dataset in the same manner. 
% sampled in a stratified manner by choosing examples from the least to highest disagreement 

% To accurately replicate the experiments, we contacted the authors of the 2022 dataset to find out more about the task design and other characteristics that were not previously released in the paper. 
The complete annotated data reflecting 500 annotations is available publicly\footnote{\url{https://github.com/XXX}}.
% \footnote{https://github.com/google-research-datasets/toxicity-replication}. 
% https://github.com/google-research-datasets/toxicity-replication
With this data release, we hope to pave the way for exploring subjectivity through an examination of the annotators' rationales as well as the expert annotator's perspectives. 
This dataset is intended to be used only for research purposes\footnote{It is especially important to note that derivatives of data accessed for research purposes should remain within the confines of research contexts and should not be employed outside of those specific settings}. 
% Besides that, it also opens a new path for understanding subjectivity by looking at the rationale of the annnotators. 
% We discuss the details of toxicity and its subtypes (derived from the prior work of Goyal et al. \cite{goyal2022your}), which are annotated in the 2023 Dataset (expert annotators)in Section~\ref{sec:task_design_and_analysis}.
We discuss the details of the data collection and analysis in Section~\ref{sec:task_design_and_analysis}.

% in detail the toxicity and toxicity sub-types ( taken from prior work \cite{goyal2022your}) that were asked to annotate in the 2023 Dataset (expert annotators)  in Section~\ref{sec:task_design_and_analysis}.

   % \end{itemize}      

\subsubsection{Data Collection and analysis}
\label{sec:task_design_and_analysis}
\begin{figure}
    \centering
    \includegraphics[width=0.45\textwidth]
    {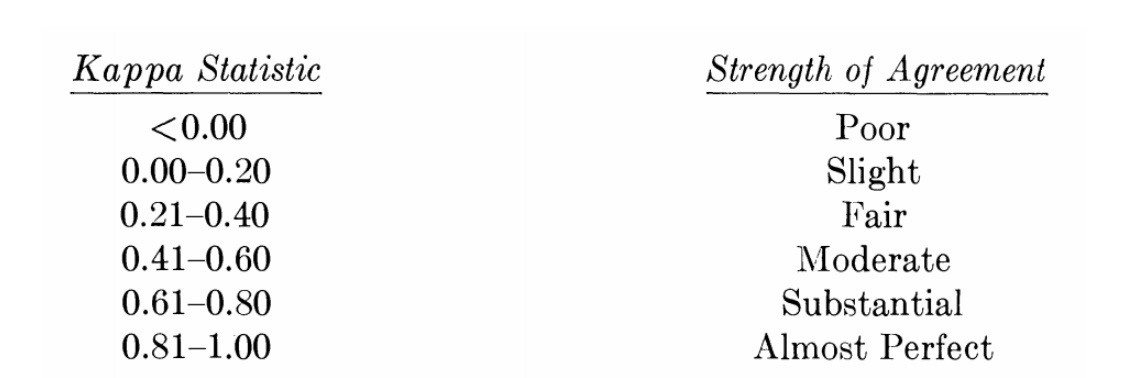}
\caption{Agreement measure interpretation for categorical data \cite{landis1977measurement}}
\label{fig:quantative_interpretation}
\end{figure}

% \subsubsection{Task Design}
Two different pools of expert annotators were presented with an identical full set of 50 comments. Details about annotations are provided in Appendix~\ref{sec:appendix_toxicity_sub_type}. 
% The data collection protocol is approved by an ethics review board.
The first pool, referred to as ``expert pool 1,'' consists of individuals with 5 or more years of experience working with subjective datasets. The second pool, referred to as ``expert pool 2,'' comprises annotators who have one year of experience working with subjective data.
% One pool has 5 or more years of experience working with subjective datasets, represented as \textit{expert pool 1} and the other pool of annotators has one year of experience working with subjective data, represented as \textit{expert pool 2}. 
We recruited six annotators from expert pool 1 and four annotators from expert pool 2. 
% In total, the 2023 dataset (expert annotators) therefore contains 10 annotations per comment. 
% Each task has instructions, and participants were urged to carefully read the instructions before beginning their annotations. 
We asked each expert annotator to rate each comment on the basis of toxicity as well as toxicity sub-type attributes. The task follows the same definition and Likert scale for each component of toxicity and its sub-types as mentioned in this paper \cite{goyal2022your} except for the new toxicity sub-type that we added. We present the toxicity and toxicity sub-types taken from prior work for completeness in Appendix~\ref{sec:appendix_toxicity_sub_type}.
% \begin{enumerate}
%     \item Toxicity is defined as ``a rude, disrespectful, or unreasonable comment that is likely to make people leave a discussion''. This is measured on a 4-point Likert scale with values between -2 and 1, where -2 = Very toxic, -1 = Toxic, 0 = Unsure, and 1 = Not toxic.
%     \item All other toxicity sub-types are mentioned in Appendix ( under Section Toxicity Subtype Definition)
%     % \item Profanity or Obscenity is defined as ``swear words, curse words, or other obscene or profane language''. This is measured on a 3-point Likert scale from -1 to 1.
%     % \item Insult is defined as ``insulting, inflammatory, or negative comment towards a person or a group of people''. This is measured on a 3-point Likert scale from -1 to 1.
%     % \item Threatening describes ``an intention to inflict pain, injury, or violence against an individual or group''. This is measured on a 3-point Likert scale from -1 to 1.
%     % \item Identity-based negativity is defined as ``negative or hateful comments targeting someone because of their identity''. This is measured on a 3-point Likert scale from -1 to 1.
% \end{enumerate}
 Besides the toxicity-related components in the 2017 and 2022 datasets, we 
 added one new component of the toxicity sub-type, as well as an open-ended response question for rationale in the task to understand the annotators' thinking styles and processes, thereby increasing transparency and data quality.
% By leveraging the insights of these experts, our approach enhances the transparency and interpretability of the toxic dataset, leading to more meaningful and reliable annotations.
 In doing this, we take inspiration from prior work \cite{mcdonnell2016relevant,sun2022investigating,liu2006annotation}. 
 We define this new toxicity-related sub-type and rationale as the following:
 % \textit{Leaving Discussion} is described as ``comments which are likely to make you leave or not want to continue a discussion''. This is measured on a 3-point Likert scale from -1 to 1.
 % \textit{Explanation}is an open-ended response that is defined as ``the rationale behind your judgment to choose the comment as toxic or not.''
 \begin{enumerate}
     \item Leaving Discussion is described as ``comments which are likely to make you leave or not want to continue a discussion''. This is measured on a 3-point Likert scale from -1 to 1.
     \item Explanation is an open-ended response that is defined as ``the rationale behind your judgment to choose the comment as toxic or not.'' \\
     % We provided some suggestions for explanations such as ``too vague/lacks context,'' ``sarcastic,'' ``accusatory,'' ``ignorant,'' ``obtuse,'' etc.
 \end{enumerate}% \subsubsection{Task Analysis}
 In this paper, to better understand the subjectivity of annotations for each dataset we first calculate the inter-rater reliability (IRR) of the 2017 dataset, 2022 dataset (subgroups: Control, LGBTQ, and African American), and 2023 dataset (expert annotators). 
 % using Krippendorff's alpha \cite{krippendorff2011computing}. 
% In this paper, we refer to this Figure 
We choose to employ inter-rater reliability (IRR) as a metric to quantify the level of agreement among annotators' toxicity annotations as this fits this specific problem. Among all existing IRR coefficients, \cite{krippendorff2011computing} stands out as 
 % it can handle more than two raters and account for missing values, 
 highly suitable for all the datasets utilized in this paper. 
 % \citet
 % All comments are first analyzed quantitatively in order to understand the disagreements between experts and non-expert annotators using inter-rater reliability (IRR). 
 Additionally, to measure the disagreement between annotators across different groups, we adopt cross-replication reliability(xRR) and normalized xRR which are based on Cohen's kappa \cite{wong2021cross}. 
 % In the case of xRR, annotators need to be taken from two different groups. 
 We measure the disagreement between all combinations of 2022 sub-groups (e.g. Control, LGBTQ and African American) and 2017 vs 2022 subgroups.
% Both these tables show disagreement between the 2017 and the 2022 subgroups. 
In addition to that, we present the disagreement metrics among annotators selected from two different groups between all combinations of 2017, 2022 Control (here we have chosen 2022 Control as a baseline (as this group consists of neither LGBTQ nor African Americans) to represent results without any bias from any specific group), as well as expert pool 1 and expert pool 2.
% in Table \ref{table:XRR3}.
% Table \ref{table:XRR1} describe the subjectivity between the 2022 subgroups and Table \ref{table:XRR2} shows the subjectivity between  the 2017 dataset and 2022 sub-groups. Finally, we present the disagreement among annotators from all three different datasets - 2017, 2022 control (here we have chosen 2022 control as a baseline to reduce biases), and 2023 dataset (expert) in Table \ref{table:XRR3}.
 % Therefore, in this paper, we employ xRR and normalized xRR to measure disagreement between annotators from distinct experiments. 
 % In order to measure the disagreement across different groups we additionally use cross-replication reliability (xRR) \cite{wong2021cross}. 
%  In this paper, we choose to employ inter-rater reliability (IRR) as a metric to quantify the level of agreement among annotators' toxicity annotations as this fits this specific problem. Among all existing IRR coefficients Krippendorff's alpha \cite{krippendorff2011computing} stands out as 
% % it can handle more than two raters and account for missing values, 
% highly suitable for all the datasets utilized in this paper. Additionally, to measure the disagreement between annotators from two different experiments, we adopt cross-replication reliability (xRR) and normalized xRR which are based on Cohen's kappa proposed by Wong et al. \cite{wong2021cross}. Therefore, in this paper, we employ xRR and normalized xRR to measure disagreement between annotators from distinct experiments.
We describe more details of dataset transformation to calculate IRR and xRR in Appendix~\ref{sec:appendix_toxicity_sub_type}. 
% In terms of the 2017 and 2022 datasets, we take into account 25,500 comments which are common in both these datasets. The 2017 dataset has a varying number of annotators for each comment whereas for the 2022 dataset each group has 5 annotators. Therefore we consider random 5 annotators for each comment from the 2017 dataset. 
%  We transform the 2023 dataset (expert annotators) dataset and the 2022 dataset to binary scale since the 2017 dataset is in binary scale in order to maintain consistency and quantitatively analyze all three datasets in an effective manner. 
 In terms of IRR, we analyze annotators' agreement for all toxicity and toxicity sub-types to show a distribution of agreement among annotators. 
 % Based on the IRR result we focus only on the toxicity attribute as this is the most subjective and the one which directly reflects toxicity across all three datasets. 
 % We select only the toxicity attributes for xRR and normalized xRR analysis based on the understanding that it is the most subjective attribute and directly represents the concept of toxicity across all three datasets. 
 We concentrate solely on the toxicity attribute, disregarding the sub-types of toxicity for both xRR and normalized xRR analysis. This decision stems from the IRR result that toxicity is the most subjective attribute, and it unequivocally symbolizes the concept of toxicity across all three studied datasets.
 % Therefore in the case of xRR, we choose only the toxicity attribute  from the above-mentioned three datasets (e.g. 2017 dataset, 2022 dataset, and 2023 dataset (expert annotators)). 
 % and also from the IRR analysis we see that toxicity has higher disagreement than other toxicity sub-types. Therefore in the case of xRR, we choose only the toxicity column  from the above-mentioned three datasets (e.g. 2017 dataset, 2022 dataset, and 2023 dataset (expert)).
 % We also quantitatively analyze all the comments which belong to high disagreement based on toxicity for two different pools as well as to high disagreement across two pools of experts for toxicity.
 % In this paper, we refer to this absolute interpretation of IRR and xRR as the Landis-Koch approach (shown in Figure \ref{fig:quantative_interpretation}) as we are using Krippendorff's alpha which is based on Cohen's kappa as well as xRR is based on Cohen's kappa and we are using categorical data for toxicity annotations.
 In this study, we adopt the Landis-Koch approach (illustrated in Figure \ref{fig:quantative_interpretation}) to refer to the absolute interpretation of IRR, xRR, and normalized xRR. 
 This choice is motivated by the choice of metric and utilization of categorical data for toxicity annotations. 
 % We also quantitatively analyze all the comments which belong to high disagreement based on toxicity for two different pools as well as to high disagreement across two pools of experts for toxicity and leaving discussion. We discuss in detail the conditions to consider a comment as highly disagreed in Section 4. 

 % In addition to quantitative analysis, we qualitatively examine the sources of subjectivity. We first collected all open-ended responses for the rationale from each expert annotator. Then we started the first round of coding after looking at all the responses carefully. We created five high-level themes using thematic analysis \cite{braun2012thematic} from rationales where expert annotators do not agree with the toxicity of the comment. We discuss in more detail the quantitative and qualitative analysis later in this section ( Section 3.2). % in Section 4.

% we created five high-level themes from rationales where expert annotators mention the comment as toxic, where they do not agree with the toxicity of the comment, and when they are not sure of the toxicity of the comment, respectively. We discuss in more detail about the quantitative and qualitative analysis later in this section. % in Section 4.

\subsection{Result}

Here we present our findings and analysis on the quantification of subjectivity using inter-rater reliability (IRR) and cross-replication reliability metric (xRR and normalized xRR).

% Finally, we discuss the sources of subjectivity using the rationale provided by expert annotators. 
\begin{table*}
\centering
\scalebox{0.86}{
\begin{tabular}{|l|c|c|c|c|c|}
\hline
\textbf{Dataset}               & \textbf{Toxicity}                               & \textbf{Profanity}                              & \textbf{Insult}                                & \textbf{Threat}                                 & \textbf{Identity Attack}                        \\ \hline
2017                  & 0.136          & 0.222          & 0.208          & 0.124          & 0.142          \\ \hline
2022-Control          & 0.240          & 0.242          & 0.209          & 0.125          & 0.182          \\ \hline
2022-LGBTQ           & 0.268 & 0.290 & 0.234 & 0.142 & 0.238 \\ \hline
2022-African American &0.202          &  0.219          & 0.184          & 0.120          & 0.157          \\ \hline
\end{tabular}}
    \caption{Calculated value of IRR for 2017 dataset, 2022 dataset  for three different groups for toxicity and toxicity sub-types}
    \label{table:IRR1}
    % \vspace{-10mm}
\end{table*}

% \subsubsection{}section{Quantative Analysis}
% \subsubsection{Disagreement between annotators}

\subsubsection{Quantitative Analysis using IRR}

% \subsubsection{Understanding subjectivity using IRR}
% To better understand the subjectivity of annotations for each dataset we first calculate the inter-rater reliability (IRR) of the 2017 dataset, 2022 dataset, and 2023 dataset (expert annotators) using Krippendorff's alpha \cite{krippendorff2011computing}. 
% In this paper, we refer to this Figure \ref{fig:quantative_interpretation} for a comprehensive understanding of how to interpret the IRR values.
% In terms of the 2017 and 2022 datasets, we take into account 25,500 comments which are common in both these datasets. The 2017 dataset has a varying number of annotators for each comment whereas for the 2022 dataset each group has 5 annotators. Therefore we consider random 5 annotators for each comment from the 2017 dataset.

% \begin{itemize}
%     \item Agreement within expert pool 1
%     \item Agreement within expert pool 2 
%     \item Agreement within 2017 
%     \item Agreement within 2022 
% \end{itemize}

% For the 2022 dataset, there are three different specialized rater pools (African American, LGBTQ, and control) we consider these three groups in order to understand the intra-group disagreements. We also calculate the IRR for all three groups together to understand the overall annotator agreement of the 2022 dataset. We use both the 2017 and 2022 datasets as a baseline for understanding the subjectivity of toxic datasets. For the 2023 dataset (expert annotators), we calculate the IRR for each pool of expert annotators.
\begin{table*}
% \scalebox{0.86}{
\centering
\begin{tabular}{|l|c|c|c|c|c|c|}
\hline
\textbf{Dataset}               & \textbf{Toxicity}                               & \textbf{Profanity}                              & \textbf{Insult}                                & \textbf{Threat}                                 & \textbf{\makecell{Identity \\Attack}}              &
\textbf{\makecell{Leaving \\ Discussion}}            \\ \hline
\makecell{Expert \\ pool 1}                  & 0.269         & 0.586          & 0.304          & 0.578         & 0.525     & 0.279        \\ \hline
\makecell{Expert \\ pool 2}         & 0.400         & 0.813         & 0.620     & 0.630         & 0.695        & 0.202 \\ \hline
% \makecell{Expert\\ All}            & \textbf{0.329} & \textbf{0.659} & \textbf{0.420} & \textbf{0.616} & \textbf{0.575} &\textbf{0.298}\\ \hline
\end{tabular}
    \caption{Calculated value of IRR for both pools of expert annotators for toxicity and toxicity sub-types}
    \label{table:IRR2}
    % \vspace{-10mm}
\end{table*}
% All three datasets (e.g. 2017, 2022, and 2023 dataset (expert annotators)) have the same columns for toxicity and toxicity sub-types:  profanity or obscenity, insult, threat, and identity attack. Along with all these columns, the 2023 dataset (expert annotators) has the new `leaving discussion' column which is not present in the other two datasets. 
% The IRR values for the 2017, 2022, and 2023 datasets (expert annotators) are presented in Table~\ref{table:IRR1} and Table~\ref{table:IRR2}. 
% These values provide insights into the level of agreement among the annotators within each dataset.

\begin{itemize}
    % \item Inter-Rater Reliability (IRR) Comparison: The IRR values for the 2017, 2022, and 2023 datasets (expert annotators) are presented in Table~\ref{table:IRR1} and Table~\ref{table:IRR2}. These values provide insights into the level of agreement among the annotators within each dataset.

   \item  \emph{Highest Disagreement dataset among three datasets}: Our analysis reveals that the 2017 dataset exhibits the highest level of disagreement (IRR = 0.136) among its annotators in terms of not only toxicity but also across all other toxicity sub-types among all three datasets including 2017, 2022 sub-groups and 2023 (expert annotators)(see Table~\ref{table:IRR1}). 
   % This finding indicates that there is substantial disagreement in the annotations provided for toxicity and toxicity sub-types within the 2017 dataset. 
   The high level of disagreement suggests that there may be diverse interpretations or subjective judgments among the annotators of 2017 dataset when labeling toxic content within this dataset.

   \item \emph{Disagreement Among Sub-Groups}: For the 2022 dataset we observe from our result that the 2022 African American group has the highest disagreement for toxicity and toxicity sub-types among all 2022 sub-groups (e.g.\ Control, LGBTQ, and African American; see Table~\ref{table:IRR1}).
   This finding sheds light on the distinct characteristics and complexities associated with the African American group's perceptions and interpretations of toxic content within the dataset.
    The result suggests that there may be varying viewpoints or subjective judgments regarding what constitutes toxicity or its sub-types within a specific demographic. 
    % These differences in interpretation could stem from cultural, social, or individual factors that influence their understanding and identification of toxic elements.

    \item \emph{Disagreement Between Expert Annotators}: In the case of the 2023 dataset (expert annotators), we observe that expert pool 1 demonstrates a higher level of disagreement among annotators compared to expert pool 2 (see Table~\ref{table:IRR2}). 
    This discrepancy highlights the influence of the specific group of expert annotators on the agreement levels within the dataset. 
    The subjectivity between the two expert pools could be attributed to differences in their expertise, individual biases, or interpretation variations.

%     \item Difference among toxicity sub-types disagreement between three datasets: 
%     % In terms of other toxicity sub-types, we see this trend in terms of disagreement Threat > Identity Attack > Insult > Profanity for both the 2017 and 2022 dataset, whereas for the 2023 dataset (expert annotators) we see the trend Insult > Identity Attack > Threat> Profanity. Also leaving discussion has the most disagreement among all toxicity sub-types.

%    In both the 2017 and 2022 datasets, the order of disagreement for toxicity sub-types is: Threat > Identity Attack > Insult > Profanity.
%    However, in the 2023 dataset with expert annotators, the trend differs slightly: Insult > Identity Attack > Threat > Profanity.
%    Also for the 2023 dataset (expert annotators) Leaving discussions generate the highest level of disagreement among all toxicity sub-types across the datasets.
% These findings highlight the varying levels of disagreement and trends observed in labeling toxicity sub-types. 
% Understanding these patterns can improve annotation guidelines and contribute to a better understanding of the complexities associated with identifying and classifying toxic content accurately. 
\end{itemize}
\subsubsection{Quantative analysis using xRR and normalized xRR}
% \subsubsection{Understanding subjectivity using xRR and normalized xRR}
\begin{table}
\centering
\scalebox{0.86}{
\begin{tabular}{|l|l|l|}
\hline
\textbf{Dataset}                      & \textbf{\makecell{Normalized \\ XRR Value}}        & \textbf{\makecell{XRR \\Value}}                    \\ \hline
\makecell{ 2022-Control vs \\ 2022-African-American} & 0.836          & 0.184          \\ \hline
\makecell{2022-Control vs\\ 2022-LGBTQ}           & 0.818          & 0.207          \\ \hline
\makecell{2022 African-American \\ vs 2022 LGBTQ}   & 0.800 & 0.186 \\ \hline
\end{tabular}}
    \caption{Cross-replication reliability using xRR and normalized xRR of toxicity for all possible combinations of 2022 sub-groups}
    \label{table:XRR1}

\end{table}

\begin{itemize}

   \item \emph{Least disagreement Group}: Our analysis using xRR and normalized xRR reveals that the two expert pools show the least disagreement compared to all other combinations. (see Table~\ref{table:XRR1}, Table~\ref{table:XRR2} and Table~\ref{table:XRR3}). This indicates that annotations from expert annotators exhibit the highest level of similarity among the three datasets across different groups.
    \item \emph{Differences Between 2017 and 2022 Datasets}: When analyzing the 2017 and 2022 datasets, we found that the 2017 dataset shows a higher disagreement with all sub-groups from the 2022 dataset compared to the disagreements observed within the various combinations of the 2022 sub-groups (Table~\ref{table:XRR1} and Table~\ref{table:XRR2}). This disparity implies that the two datasets may differ significantly in terms of their characteristics or underlying factors influencing the subjective annotations.
    % \item Highest agreement among all combinations: Our analysis using xRR and normalized xRR reveals that the two expert pools show the least disagreement compared to all other combinations. (see Table \ref{table:XRR1}, Table \ref{table:XRR2} and Table \ref{table:XRR3}). This indicates that annotations from expert annotators exhibit the highest level of similarity among the three datasets across different groups.
    \item \emph{Among Expert Pools}: 
    In both the 2017 dataset and the 2022 control group, we observed that expert pool 1 exhibits a higher level of disagreement compared to expert pool 2 (Table~\ref{table:XRR3}). This discrepancy within the expert pools suggests that even among experts, individual differences and subjective perspectives can influence the annotation process, leading to varying levels of agreement.
     \item \emph{Role of annotator's background}: Notably, our results suggest that the 2023 dataset (expert annotators) is more similar to the 2017 dataset than the 2022 Control group. This implies that annotators' background has a more significant impact on subjectivity than the timeline when considering these three datasets (2017, 2022, and 2023(expert annotators)).

\end{itemize}

% These findings provide insights into the agreement levels among different groups of expert  and non-expert annotators, the differences in disagreement within datasets, and the influence of annotators' backgrounds on subjectivity. Understanding these dynamics shed light on the subjective nature of the analyzed datasets.
% Our result using xRR and normalized xRR shows that the two expert pools have the lowest disagreement among all the combinations of expert pools with the 2017 and 2022 control group.
% This shows that annotations between expert annotators are the most similar among all three datasets (see Table \ref{table:XRR3}). 
% In terms of the 2017 and 2022 datasets, the 2017 dataset has a higher disagreement with all 2022 sub-groups than among all combinations of 2022 sub-groups(see Table \ref{table:XRR1} and \ref{table:XRR2}). Expert pool 2 has a higher disagreement compared to expert pool 1 across both the 2017 dataset and 2022 control dataset (see Table \ref{table:XRR3}). 

% Figure \ref{fig:quantative_interpretation} serves as a crucial point of reference for comprehending the interpretation of xRR and normalized xRR values.
% Above all, the most interesting part of our result suggests that the 2023 dataset (expert annotators) is more similar to the 2017 dataset than the 2022 dataset. This clearly indicates that the annotators' background plays a more important role than the timeline in terms of the subjectivity of these three datasets (e.g. 2017, 2022 and 2023 (expert annotator).
\begin{table}
\centering
\scalebox{0.86}{
\begin{tabular}{|l|l|l|}
\hline
\textbf{Dataset}              & \textbf{\makecell{Normalized\\ XRR Value}}         & \textbf{\makecell{XRR\\ Value} }                   \\ \hline
\makecell{2017 vs\\ 2022-Control}         & 0.648          & 0.117          \\ \hline
\makecell{2017 vs \\ 2022-\\ African-American} & 0.664          & 0.110          \\ \hline
\makecell{2017 vs \\ 2022-LGBTQ} & 0.570 & 0.109 \\ \hline
\end{tabular}}
    \caption{Cross-replication reliability using xRR  and normalized xRR of toxicity for all possible combinations between 2017 and 2022 sub-groups (e,g, Control, African American and LGBTQ).}
    \label{table:XRR2}
\end{table}

% Overall our quantitative analysis indicates not only non-expert annotators but also expert annotators exhibit a major disagreement for toxicity among raters and across different pools. As a result, it is very important to understand the reasons behind the disagreement.

Our overall findings from the quantitative analysis show that there is  a lot of subjectivity in toxic datasets even among expert annotators. Our result shows that there are high levels of disagreement in toxicity annotations between annotators who are on the same team ranging the IRR value from 0.1 to 0.4.  
% A considerably more disagreement is found from the result  among expert annotators from the same team over what causes people to leave a discussion ($0.20 < IRR < 0.28$). 
The disagreement is not only limited to the same team; it also affects several groups of annotators. We see a range of toxicity disagreement across different groups of annotators and time, with normalized xRR values between 0.1 and 1 (xRR values between 0.04 and 0.3).
Overall the result demonstrates that not only annotators in general, but even expert annotators have a high disagreement of toxicity annotation ($0.200 < IRR <= 0.400$).
% and also there is more disagreement in one pool of expert annotators than other pools of expert annotators.
This implies that each annotator contributes a different perspective when annotating toxicity-related comments.
% We then show the model of the subjectivity of diverse viewpoints by presenting an illustration within the framework of an LLM focused on personalizing the subjectivity of toxicity in Section~\ref{sec:personalizing_subjectivity_llm}.
We then use an LLM framework to demonstrate how we model the subjectivity of toxicity. Details are shown in Section~\ref{sec:personalizing_subjectivity_llm}.
% As a result, it is very important to understand the reasons behind the disagreement.

\begin{table*} 
% \scalebox{0.88}{
\centering
\begin{tabular}{|l|l|l|}
\hline
\textbf{Dataset}               & \textbf{\makecell{Normalized \\ XRR Value}}         & \textbf{\makecell{XRR \\ Value}}                    \\ \hline
\makecell{Expert pool 1 vs\\ Expert pool 2} & 1.073          & 0.351          \\ \hline
\makecell{2017 vs\\ Expert pool 2}          & 0.846          & 0.197          \\ \hline
\makecell{2017 vs \\ Expert pool 1 }         & 0.728 & 0.139 \\ \hline
\makecell{2022-Control vs \\ Expert pool 2} & 0.415                                  & 0.128                                  \\ \hline
\makecell{2022-Control vs\\ Expert pool 1}  & 0.185                                  & 0.047                                  \\ \hline
\end{tabular}
    \caption{Cross-replication reliability using xRR and normalized xRR among two expert pools, among all combinations between 2017 dataset and expert pools, all combinations between 2022 Control group and expert pools}
    \label{table:XRR3}
\end{table*}

\section{Personalizing subjectivity of toxicity annotation using LLM}
\label{sec:personalizing_subjectivity_llm}

In this section, we model subjectivity further by conducting experiments using an LLM to assess its ability to mimic an individual's response to toxicity. 
% Through a series of experiments, we aim to replicate annotators' toxicity perspectives using the LLM. 
Our experimental design is motivated by a simulation-based study that examines the structure of disagreement in toxicity-related comments, which we discuss in the following section.% Real datasets with raters and items are composed of many hidden attributes  such as item properties, raters skill, bias etc. We used a clustering approach called spectral biclustering in order to disentangle these properties through clustering of rater and items.
% \subsection{Procedure}
\subsection{Motivation: Simulated Toxic Dataset Experiment}
% \label{sec:motivation}
 Real datasets that involve both annotators and comments often contain numerous underlying attributes, such as the properties of the comment, the skill levels of the annotators, and various forms of biases. 
 To disentangle these factors and gain insight into the data, we constructed a simulated dataset where the attributes are known in order to identify key patterns and relationships within the simulated data. 
 % Appendix \ref{sec:appendix_simulated} describes how we created the simulated dataset.
 % To simulate a dataset, we first create an entire matrix of ratings from $N$ annotators and $M$ items, resulting in a fully rated dataset. We then reduce the replication to more realistic levels by randomly sampling down ratings per item. If the replication is $P$ that refers to $M$ items annotated by $P$ annotators.
 % We used Weighted Alternating Least Squares (WALS) to identify key patterns and relationships within the data. 
 % We discuss the experiment in detail below.

 \subsubsection{Method}
 In this experiment, we first simulated a dataset using different attributes for comment and annotator (details of how we have simulated the dataset is in Appendix \ref{sec:appendix_simulated}).
 % In this experiment, to simulate a dataset, we first create an entire matrix of ratings from $N$ annotators and $M$ items, resulting in a fully rated dataset. We then reduce the replication to more realistic levels by randomly sampling down ratings per item. If the replication is $P$ that refers to $M$ items annotated by $P$ annotators. 
 % Then we establish a set of attributes(e.g.\ annotator skill level and comment difficulty level). For annotators' skill level, there are 4 attributes: (1) Expert (2) Average (3) Bad (4) Random whereas in the case of comment difficulty level, there are 3 attributes : (1) Easy (2) Normal (3) Hard. The details are described in Appendix \ref{sec:appendix_simulated}). We randomly initialize attributes to each annotator and item and use these attributes to probabilistically assign an annotation for each annotator and item interaction. We then use the WALS factorization approach \cite{hu2008collaborative} to create items and annotator embeddings (see Appendix \ref{sec:appendix_simulated} for the details) and finally we use HBDSCAN for clustering that.
 We used a total pool of 500 annotators and 5000 comments. 
 The ratio to simulate ground truth, annotator type, and comment type in each replication is described in Appendix~\ref{sec:appendix_simulated}. 
 We started the experiment with the full dataset comprising 500 annotators and 5000 comments. 
Using WALS factorization \cite{hu2008collaborative}
 with dimension size 3, regularization of 0.1, and 5 iterations, we generated annotator and item embeddings with an associated error (refer to Appendix~\ref{sec:appendix_simulated}).
Subsequently, we conducted 200 replications for 5000 items, randomly selecting 200 annotators out of a total of 500 to annotate each of the 5000 comments, while maintaining the same error as described above. 
% The intentionally higher annotator value allowed us to examine the model's impact on understanding the differences in annotation structure.
Furthermore, we examined the influence of lower replication sizes (100, 50, 20, 15, 10, and 5) under various settings (see Appendix~\ref{sec:appendix_simulated}). Then we clustered using HDBSCAN.
\begin{figure}
    \centering
    \includegraphics[width=0.70\textwidth]
    {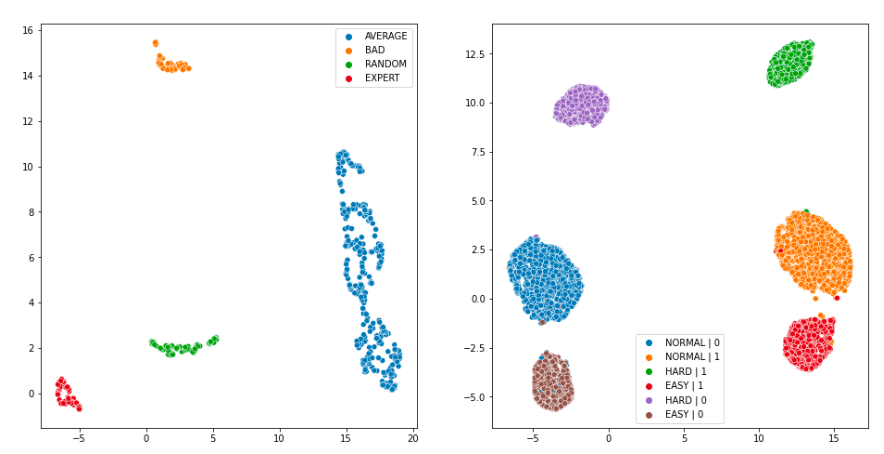}
\caption{UMAP Projection of annotator and items using full simulated data.}
\label{fig:full_data}
\end{figure}

\subsubsection{Result}
% The embedding of annotators and items in the full dataset using UMAP projection demonstrates the separation of annotators and objects quite effectively(see Figure \ref{fig:full_data}).
% % ** Note: Figure 8 add x = false Positive rate, y = True Positive rate in the figure
% The replication of 200, employing the identical setup as the full dataset, exhibits a similar distinction between annotator and item spaces, as observed in the experiment with the complete dataset (Figure \ref{fig:full_data}).
% However, with smaller replication sizes, there is no significant differentiation observed between item and annotator space.
% Furthermore, when utilizing a majority vote as the ground truth, the AUC decreases as the replication size decreases (refer to Appendix \ref{sec:appendix_simulated} for details).

\begin{figure}
    \centering
    \includegraphics[width=0.70\textwidth]
    {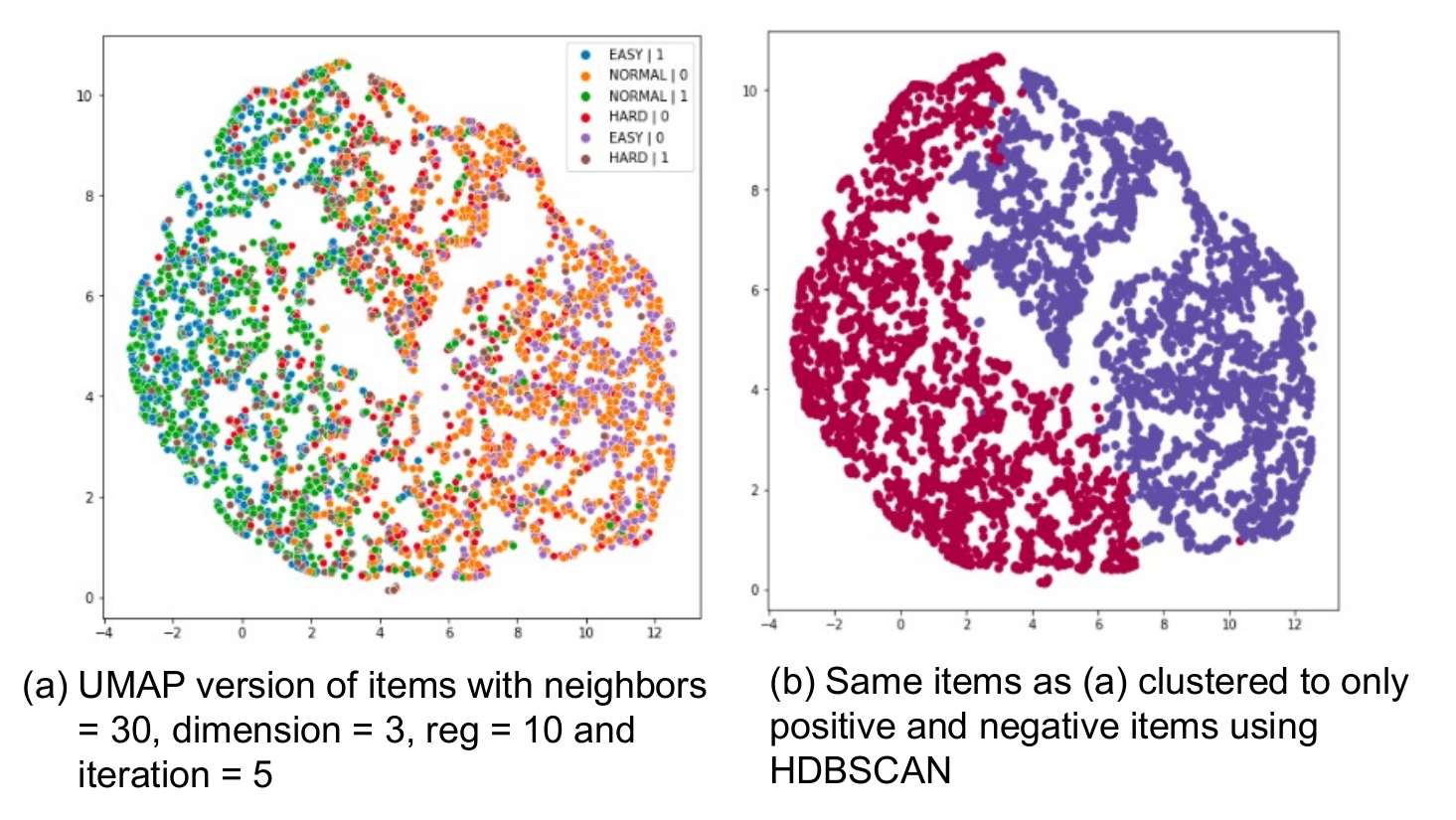}
\caption{Replication 5 of items}
\label{fig:replication5}
\end{figure}

\begin{figure}
    \centering
    \includegraphics[width=0.55\textwidth]
    {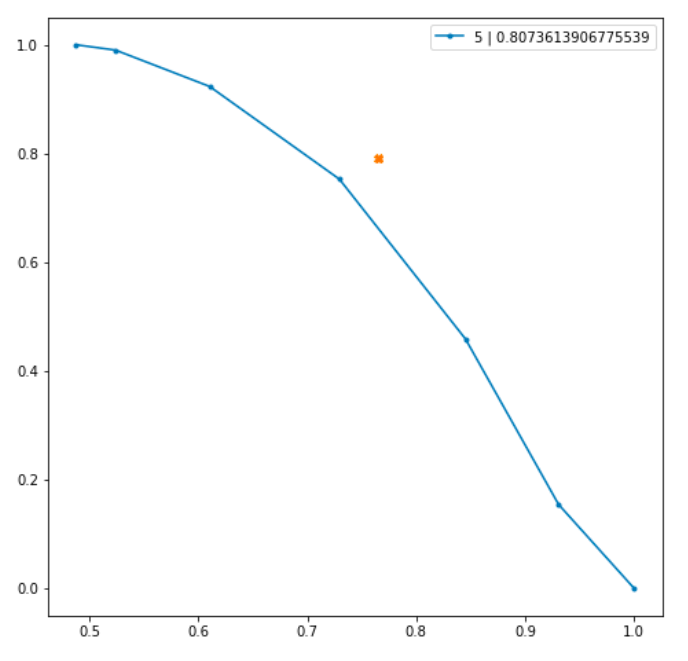}
\caption{PR Curve of Replication 5 using majority as a ground truth and using binary level as a ground truth.}
\label{fig:replication5_prcurve}
\end{figure}

UMAP projection of the full dataset effectively separates annotators and items (Figure~\ref{fig:full_data}). The replication with 200 data points replicates this distinction. However, smaller replication sizes show no significant differentiation between item and annotator space. 
Exploring different options (see Appendix~\ref{sec:appendix_simulated}), we found that lower replication results in improved separation between annotators and items (Figure~\ref{fig:replication5}). These findings suggest that exposing the model to actual data, rather than average data per item, improves performance with small replication. 
Additionally, we observed from our result when using a majority vote as the ground truth, AUC decreases as replication size decreases (see Appendix~\ref{sec:appendix_simulated} for details). However using the binary label from HDBSCAN as ground truth achieves higher AUC compared to majority voting, as shown in Figure~\ref{fig:replication5_prcurve}. Our result shows that the model benefits from accessing raw annotations rather than relying on majority voting in order to better learn the separation between comments and annotators for small replication.
% Replication 200 using the same setup  of the full dataset demonstrates the same separation of both annotator and item space as shown from the experiment using the full dataset. (as shown in Figure \ref{fig:full_data}).
% However, lower replication sizes  do not demonstrate a significant distinction between item groups and annotator groups. 
% When a majority vote is used as ground truth then the AUC decreases as the replication decreases (see Appendix \ref{sec:appendix_simulated}).
% the impacts of employing a majority vote as a proxy ground truth on the AUC by replication size.
% Figure \ref{fig:different_replication_effect} 
% AUC decreases as replication decreases.
% We used various options (described in the \ref{sec:appendix_simulated}) in the experiment to make improvements with lower replication.
Overall our findings from the simulated experiment indicate that  distinguishing between expert, bad, random, and average annotators is crucial for enhancing the model's performance.

% Overall our findings from the simulated experiment indicate that exposing the model to actual data, rather than average data per item, improves performance with small replication. Additionally, distinguishing between expert, bad, random, and average annotators is crucial for enhancing the model's performance. 
% Furthermore, our results demonstrate that the model benefits from accessing raw annotations rather than relying on majority voting for comments and annotators.

% Overall, our findings suggest that the model needs to be exposed to actual data rather than average data per item in order to perform better with small replication.
% Apart from that, it highlights the importance of distinguishing between expert, bad, random, and average annotators in order for the model to perform better.
% In addition to that, our result shows that the model will be able to perform better if the model can see raw annotation rather than the majority voting for comment and annotator.
% that the higher the replication there is very clear separation in both rater and item space. 
% We lost a lot of separation power for clustering from replication 50 to replication 20.

\subsection{LLM Experiment: Can subjectivity be personalized? }
% The result from simulated dataset demonstrate that model performs better with expert raters as they have a strong pattern and logical relationship as well as as the result shows that model needs to see each rater's annotation apart from the majority annotation in order to perfoem better.
The goal of the study is to understand if an LLM can mimic annotators' viewpoints for toxicity annotations.

% In order to achieve this goal, we first determined the number of training examples required for the model to mimic an annotator. Thereafter we tested the model's performance of imitation for testing with the 2017 dataset whose training split is used for training the model and also testing with the 2023 dataset (expert).

% Based on the result of training examples, we subsequently identified the cosine similarity between the model generated explanation with expert rater's explanations to determine whether the model can generate explanation similar to those provided by the expert raters from 2023 dataset.

% \subsubsection{Model Training}

% We train our LLM using soft prompt tuning ~\cite{lester-etal-2021-power}. This allows us to efficiently tune the LLM on 10 - 500 examples. The LLM we use is the 62b FLAN-cont-PaLM ~\cite{flan_palm}. We tuned a prompt consisting of 100 tokens, each with an embedding of dimension 8192 for the 62B model. We trained these tokens with a basic Adam optimizer with clipped gradients. We chose a default learning rate and schedule for all our experiments and did not tune hyperparameters further. We set the sampling temperature to 0 and fixed the prompt initialization, to reduce randomness. For each training run, we have a train, validation, and test set with balanced positive and negative classes. We train for 20-40 epochs and use a validation set to pick the final model for each run.
\subsubsection{Method}
We discuss the model training in Appendix~\ref{sec:appendix_model_training}.
We employ two approaches in this case: varying training sizes and two sets of annotators as test sets.
In the first approach, we manipulated the size of the training data to examine how it influences LLM's ability to learn and comprehend the unique perspectives of annotators regarding toxicity. 
By changing the training data size, we can understand the model's capacity to effectively capture individual annotator viewpoints.
To train the LLM, we choose training example sizes of 50, 100, 250, and 400 from a pool of 10 annotators who annotated an equal proportion of toxic and non-toxic comments. In this paper, we address these types of annotators as \textit{balanced annotators}.
We conducted the training in two different ways: (1) the model when prompted using few-shot examples \cite{brown2020language} and (2) the model when trained via soft-tuning \cite{lester-etal-2021-power}. For soft-tuning, we treat toxicity classification as a binary Yes/No token prediction problem and train with cross-entropy loss.
Furthermore, we utilized two sets of annotators for the test set. The first set consists of annotators used during model training, allowing us to evaluate the model's performance in reproducing the perspectives it was exposed to during training.  The second set comprises a completely new group of annotators, enabling us to assess the model's generalization abilities by evaluating its understanding and adaptation to previously unseen perspectives on toxicity.
We first tested the trained model (trained with the 2017 train set from 10 balanced annotators) using the 2017 dataset test split and then with the 2023 dataset (expert annotators).
% To train the LLM, we choose training example sizes of 50, 100, 250, and 400 from a pool of 10 annotators who have annotated an equal proportion of toxic and non-toxic comments. In this paper, we address these types of annotators as \textit{balanced annotators}.
% Thereafter to evaluate the effect of imitating, we first tested the trained model (trained with the 2017 train set from 10 balanced annotators) with the 2017 dataset test set from the 10 balanced annotators and then with the 2023 dataset (expert annotators) where each comment is annotated by 10 expert annotators.
% We conducted the training in two different ways: (1) the model is prompted using few-shot examples \cite{brown2020language} and (2) the model is trained via soft-tuning \cite{lester-etal-2021-power}. For soft-tuning, we treat toxicity classification as a binary Yes/No token prediction problem and train with cross-entropy loss.
% Then we use IRR \cite{krippendorff2011computing} to determine the similarity between the model's response for a comment and the annotators' response for a comment in the context of toxicity.
Finally, we evaluate the model's performance using the formula 
\begin{equation}
\makecell{
    \text{$\Delta{IRR}$} = \text{IRR with soft-tuning\footnote{IRR with soft-tuning assesses the agreement between the annotator's and the model's toxicity responses for each comment, specifically when the model is prompted via soft tuning}} - \\ \text{IRR with few shots\footnote{IRR with few shots evaluates the agreement between the annotator's and the model's toxicity response for each comment where the model is prompted using few-shot examples.}}}
\end{equation}
In order to gauge the agreement between annotators and the synthetic output generated by the model, we take into account IRR and employ Krippendorff's alpha \cite{krippendorff2011computing} as the IRR metric. 
% IRR with soft-tuning assesses the agreement between the annotator's and the model's toxicity responses for each comment, specifically when the model is prompted via soft tuning, whereas IRR with few shots evaluates the agreement between the annotator's and the model's toxicity response for each comment where the model is prompted using few-shot examples. 
The delta between the two IRR values provides the amount of increase in the ability to mimic the annotator.
\subsubsection{Result}
\begin{figure}
    \centering
    \includegraphics[width=0.80\textwidth]
    {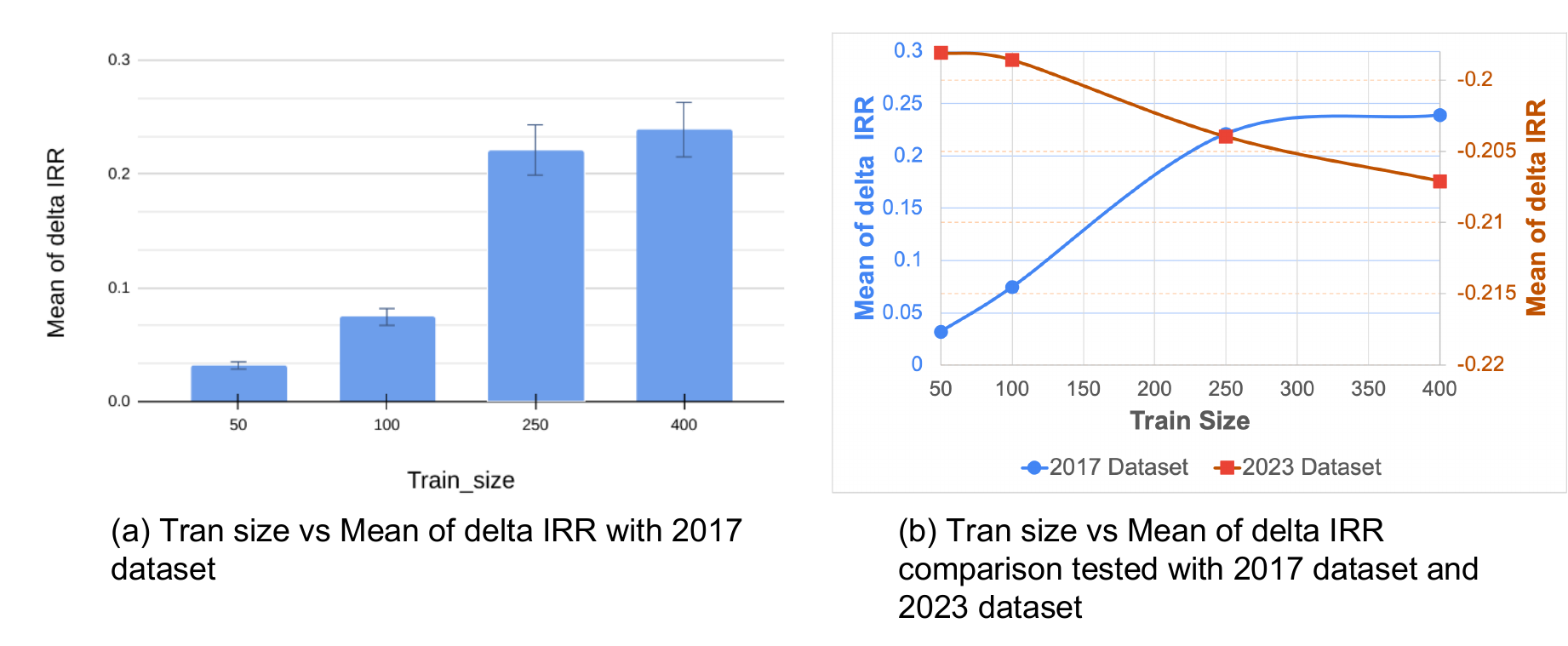}
\caption{Train dataset size vs mean of $\Delta{IRR}$.}
\label{fig:mimic_accuracy_train_size}
\end{figure}

% Here we describe model performance with respect to training size and testing set of annotators.
\textbf{Model performance w.r.t training size}: Our result (Figure~\ref{fig:mimic_accuracy_train_size} (a)) shows that the model trained with toxicity annotations performs better with more training examples (e.g. model with 400 training data performs better than the model trained with 250 training data). 
This suggests that an increase in training examples helps the model to learn each annotator's perspective better and improves the IRR. It also indicates that it will be challenging for the model to learn an annotator's toxicity annotation perspective if we have fewer than $\sim\!\!$ 100 examples. 
\textbf{Model performance w.r.t test set}:
% \item  Testing Mimicking Effect with two different sets of annotators: 
% To evaluate the effect of imitating, we first tested the trained model (trained with the 2017 train set from 10 balanced raters) with the 2017 dataset test set from the 10 balanced raters and then with the 2023 dataset, where each comment is annotated by 10 experts. 
% We similarly calculated the delta IRR as mentioned above under \textit{Training Size vs LLM accuracy}.  In terms of the 2023 dataset, we calculated delta IRR for each of the ten balanced raters, as well as the model's prediction for 10 expert raters, and averaged the delta IRR. 
Figure~\ref{fig:mimic_accuracy_train_size} (b) indicates that for the 2023 dataset (expert annotators), the smaller the training size, the better the delta IRR which is the opposite when the 2017 test split is used as a test set.
% This happens as the model is trained with the 2017 dataset using  10 balanced raters, the smaller the training size the less the model can understand each rater's perspective  from the 2017 dataset of 10 balanced raters. However, as the training size increases, more dataset from every 10 balanced annotators from the 2017 dataset is fed to the model, and the model can understand each 2017 annotator's perspective for annotation much better, which is why it shows an inverse effect with 2023 annotators. 
This occurs because the model is trained on the 2017 dataset with 10 balanced raters. As the training size decreases, the model's ability to comprehend each annotator's perspective from the 2017 dataset diminishes. Conversely, as the training size increases, the model is exposed to more data from the 10 annotators in the 2017 dataset, allowing for a better understanding of those annotators' perspectives but a poorer understanding of annotators from the 2023 dataset.

\section{Conclusion}
\label{sec:conclusion}
% In this paper, we investigated quantitatively the subjectivity of toxicity using three datasets  (three versions of CCTK dataset across different timelines) using IRR and xRR. Here 2023 version of the CCTK dataset (named as 2023 dataset (expert)) we release consisting of annotations from 10 expert annotators along with their rationale.  Due to the difficulty of sourcing experts, this dataset is limited to only 10 expert annotators. 
% In this study, we quantitatively examined the subjectivity of toxicity across different timelines using three datasets: three versions of the CCTK dataset. Notably, we introduce the 2023 dataset (expert annotators), derived from the CCTK dataset, which includes annotations and rationales from expert annotators. However, due to the challenges associated with sourcing experts, this dataset is constrained to only 10 expert annotators.
In this study, we have conducted a quantitative analysis of toxicity (used IRR, xRR, and normalized XRR) to understand subjectivity by using three datasets. Notably, we have created the 2023 dataset (expert annotators) which includes annotations and rationales from 10 expert annotators. 
% However, the limited availability of experts posed challenges in creating this dataset.
% We have conducted quantitative analysis (using IRR and xRR) within the same group of annotators (expert  and non-expert annotators) also across different groups of annotators.  
% Our result shows that subjectivity exists not only within the non-expert but also among expert annotators also across different groups of annotators and throughout different time periods.
% % Our qualitative analysis of the rationale provided by the expert annotators for their toxicity judgments from 50 comments indicates five different sources of subjectivity for the 2023 dataset (expert). They include lack of context, political comment, sarcasm, personal opinion, and when annotators are unsure about the meaning of the comment. 
Building upon our understanding of subjectivity we model subjectivity by conducting LLM experiments. This reflects the model's ability to learn annotators' diverse perspectives for toxicity annotations using two approaches: varying training data size and using two different sets of annotators as the test set.
% where one test set is set of annotators is used during model training and the other one is a completely new set of annotators.
% Our result shows that the more training example the model learns better to mimic the annotator also when the model is trained with a set of annotators it performs better when testing is done with respect to the trained annotator's annotations rather than a completely new set of annotators (see Figure~\ref{fig:mimic_accuracy_train_size}). 
% One of the limitations we had due to the small size, we could not train the model using the 2023 dataset (expert annotators).
% Overall our work suggests that disagreement not only exists among non-expert annotators but also among expert annotators and across different groups of annotators and there are many underlying factors behind subjectivity. Therefore rather than decreasing disagreement among annotators, we should understand subjectivity and employ subjective annotations as ground truth labels for the model to reflect the detection of toxicity for diverse communities. 
Our findings highlight the presence of disagreement among both non-expert and expert annotators also across different annotator groups. This indicates that subjectivity is inherent in the task of toxicity annotation. Consequently, rather than striving to reduce annotator disagreement, it is crucial to understand subjectivity. By embracing subjective annotations as ground truth labels, we can effectively capture the detection of toxicity in diverse communities, enabling models to account for the varying perspectives encountered in real-world scenarios.
When it comes to modeling subjectivity, enhancing the model's ability to learn diverse perspectives of toxicity annotations can be achieved by increasing the training data size and employing the same set of annotators as the test set used during model training rather than using a completely new set of annotators as the test set.

\section{Limitations, Ethical Considerations, and Broader Impacts}
\label{sec:ethical}

% Despite the promising result of our work to understand subjectivity using quantitative analysis as well as modeling subjectivity of toxicity using LLM, there are several limitations, ethical considerations, and broader impacts of our work which are listed below.\\
Although our work has yielded promising results in terms of understanding subjectivity through quantitative analysis and modeling the subjectivity of toxicity using an LLM, it is important to acknowledge the limitations, ethical considerations, and broader impacts associated with our research. These are outlined below:

First, in this work, we publish a dataset of toxicity annotations which are annotated by expert annotators. 
However, due to the difficulty of sourcing experts, this dataset is limited to only 10 expert annotators.
Future studies should involve more expert annotators to investigate if the subjectivity and sources of subjectivity hold true for larger numbers of experts.
In addition, considering annotated time for annotators 2023 dataset (expert annotators) is limited to only 50 comments.
The restricted dataset size prevents the utilization of the 2023 dataset (expert annotators) for training, limiting its role to that of a test set. To enhance future research, we advocate for the expansion of the toxicity dataset by incorporating additional comments. This would enable exploration of how the model can effectively learn and reflect the perspectives of expert annotators through their annotations during training.
In addition to that, we only focused on the toxicity attribute from three datasets in this paper in terms of modeling and understanding subjectivity across different groups. We suggest understanding the subjectivity of toxicity sub-types and modeling that in order to better understand the complexities associated with identifying and classifying toxic content. \\
% Due to the limited size of the dataset, the 2023 dataset (expert annotators) is employed only as a test set. 
% We encourage future researchers to expand the toxicity dataset by including more comments. This provides room for understanding how the model can learn the expert annotator's perspectives while trained with the expert annotator's annotations.
% Authors are encouraged to devote a section of their paper to concerns about the ethical impact of the work and to a discussion of broader impacts of the work, which will be taken into account in the review process
% Our research aims to understand subjectivity in toxic comments through quantitative and qualitative analysis, which has ethical implications that must be considered. By applying quantitative and qualitative methods to analyze subjective aspects of toxic comments, we strive to gain insights into the patterns and characteristics of this discourse. However, it is essential to approach this analysis ethically, taking into account privacy concerns, biases in the dataset, user consent, responsible use of AI, and the potential impact on individuals and society. We recognize the importance of maintaining privacy, addressing biases, obtaining informed consent, promoting responsible AI practices, and considering the social impact of our findings to ensure the ethicality of our research approach.
One of the important ethical implications is 2023 dataset (expert annotators) is obtained by sampling the existing 2017 and 2022 datasets in this paper, complying with the terms of use of each of these datasets. The dataset is anonymized, meaning that it doesn't include any usernames, pool identifiers, or demographic information of the individuals in the published dataset as well as in the data which we use to train our model. 
%\sherol{
Along with responsible dataset use and anonymization of participants, we also consider the broader consequences of this work and acknowledge the sensitivities of subjective spaces like toxicity. For example, while combating bias is important work towards algorithmic fairness, bias is often only implicitly represented in our data. Downstream, we can observe bias in model outputs but are less able to automate the detection of bias as a result of not having explicit demographic attributes. Practitioners and professionals have identified challenges of obtaining demographic data, but also the benefits of it \cite{andrus2021we}. While objective functions aim to generalize, we hope to see this work going beyond the average of consensus towards multi-objectives that reflect a diversity of values. %} 

One potential risk associated with this work is 
there is a risk of inadvertently amplifying or perpetuating biases present in the published dataset annotations, which may lead to biased content moderation or unintended consequences in cases where the dataset is used for research purposes.
It is essential to be mindful of these potential risks and carefully use the dataset.
% One potential risk associated with this work is 
% % the reliance on subjective annotations by ``expert annotators'' to identify toxic comments. The subjectivity of these annotations introduces the possibility of bias and variations in interpretations, which can impact the accuracy and fairness of the automated systems developed based on them. Furthermore, there is a risk of inadvertently amplifying or perpetuating biases present in the annotations, which may lead to biased content moderation or unintended consequences in real-world applications.
% % Additionally, 
% by using different sets of annotators for training and testing the model, there is a risk of introducing inconsistencies and discrepancies in capturing diverse perspectives on toxicity. The effectiveness and generalizability of the model might be compromised if it fails to accurately capture the nuances and variations in toxicity annotations from different annotator groups.
% It is essential to be mindful of these potential risks and take appropriate measures to mitigate biases, ensure fairness, and enhance the performance of automated systems for toxicity detection in diverse communities.

Finally, while our work is focused only on toxicity, it offers valuable contributions to both the understanding and modeling of annotators' perspectives in this area. Furthermore, these insights have broader implications that extend to other subjective tasks in diverse domains. These domains include sentiment analysis, irony detection, emotion detection, affect modeling, and more~\cite{pang2004sentimental,liu2010sentiment,reyes2011mining,hirschberg2003experiments,mihalcea2006corpus,alm2008affect,liu2003model}. 
% but also other subjective tasks in diverse domains, such as sentiment analysis \cite{pang2004sentimental,liu2010sentiment}, irony detection \cite{reyes2011mining},  emotion detection \cite{hirschberg2003experiments, mihalcea2006corpus}, affect modeling \cite{alm2008affect,liu2003model}, and more. 
Our work serves as a valuable resource for researchers and practitioners interested in studying and modeling subjective ground truth-based tasks. 
% It offers insights that can enhance their understanding and aid in the development of more accurate and inclusive computational models.

% Example In this paper, we demonstrate the potential threat590
% of textual backdoor attacks by showing the exis-591
% tence of a backdoor attack that is both effective and592
% stealthy. Our goal is to help NLP practitioners be593
% more cautious about the usage of untrusted train-594
% ing data and stimulate more relevant research in595
% mitigating the backdoor attack threat.596
% While an adversary may want to use our pro-597
% posed method for attacks, there are many obstacles598
% that prevent our proposed method from being harm-599
% ful in real-world scenarios. First, our threat model600
% requires the adversary to have full knowledge about601
% the training set and can control a subset. The ad-602
% versary also needs to be able to interact with the603
% trained model after it’s deployed. The constraints604
% on the threat model limit the possible scenarios605
% for our attack to be performed. Second, our pro-606
% posed attack only applies to the single sentence607
% classification task and cannot be straightforwardly608
% extended to other widely-used task formats (e.g.,609
% generation, sequence labeling, sentence pair classi-610
% fication). The constraint on the task format limits611
% its harm to real-world NLP systems beyond text612
% classification. Third, we propose a method for de-613
% fending against the attack, which can further help614
% to minimize the potential harm.

\bibliographystyle{unsrt}  
\bibliography{anthology,custom}

% %%% Comment out this section when you \bibliography{references} is enabled.

% George Kour and Raid Saabne.
% \newblock Real-time segmentation of on-line handwritten arabic script.
% \newblock In {\em Frontiers in Handwriting Recognition (ICFHR), 2014 14th
%   International Conference on}, pages 417--422. IEEE, 2014.

% George Kour and Raid Saabne.
% \newblock Fast classification of handwritten on-line arabic characters.
% \newblock In {\em Soft Computing and Pattern Recognition (SoCPaR), 2014 6th
%   International Conference of}, pages 312--318. IEEE, 2014.

% \bibitem{hadash2018estimate}
% Guy Hadash, Einat Kermany, Boaz Carmeli, Ofer Lavi, George Kour, and Alon
% \newblock Estimate and replace: A novel approach to integrating deep neural
%   networks with existing applications.
% \newblock {\em arXiv preprint arXiv:1804.09028}, 2018.

% \end{thebibliography}
\appendix
\section{Public Dataset Details}
\label{sec:appendix_dataset}

\begin{itemize}
      \item 2017 dataset: The Civil Comments Dataset consists of approximately 2 million public comments from no longer active commenting platforms. Each comment has crowdsourced labels for  toxicity and toxicity subtypes such as obscene, threat, insult, identity attack, and sexually explicit. To obtain toxicity labels, each comment was shown to up to 10 annotators who were asked to rate the toxicity of each comment. Some comments were annotated by more than 10 annotators. Annotators were also asked to indicate the toxicity sub-types for each comment. This dataset is publicly available and data itself is released under CC0 license.\footnote{\url{https://www.kaggle.com/competitions/jigsaw-unintended-bias-in-toxicity-classification/data}} This dataset uses the following Likert scale for toxicity and its sub-types:
\begin{enumerate}
  \item Toxicity is measured on a binary scale from where 0 corresponds to Not Toxic or Unsure and 1 corresponds to Toxic or Very Toxic -2 to 1 where -2 = Very Toxic, -1= Toxic, 0 = Hard to Say, and 1 = Not Toxic.
    \item Toxicity sub-type (e.g. identity attack, insult, obscene, threat) is measured again on a binary scale where 0 corresponds to Not Toxic or Unsure, and 1 corresponds to Toxic, according to the definition of that sub-type. 3-point Likert scale, where -1 = Yes, 0 = Not Sure, 1 = No.
\end{enumerate}
 \item 2022 Dataset: The Specialized Rater Pools Dataset consists of a total of 25,500 comments from the Civil Comments dataset. The authors of the 2022 dataset worked with two groups of annotators (``specialized rater pools'') from different identity groups, one LGBTQ and one African-American, and one control group from the general US population. They then sampled 8500 comments pertaining to each of the identity groups, measured by the identity labels of each group (control group comments reference neither identity).
 All three rater groups received the identical entire collection of 25,500 comments, with 5 annotators from each group allowed to annotate each comment. As a result, in total the complete annotated data including all individual annotations  consists of a total of 382,500 annotations. The dataset is publicly available on Kaggle in CSV and TSV formats as the Jigsaw Specialized Rater Pools Dataset\footnote{\url{ https://www.kaggle.com/datasets/google/jigsaw-specialized-rater-pools-dataset}}. This dataset uses a 4-point Likert scale for toxicity, where -2 = Very Toxic, -1 = Toxic, 0 = Unsure, and 1 = Not Toxic, and a 3-point Likert scale for toxicity sub-types, where -1 = Toxic (according to that sub-type), 0 = Unsure, and 1 = Not Toxic.

\section{Data Collection and analysis details}
\label{sec:appendix_toxicity_sub_type}

\subsection{2023 dataset (expert annotators) Annotator Details}
% We recruited participants by emailing them. Among the 10 participants, there are 4 female and 6 male participants. 
% We compensated the annotators by sharing our results with them which was helpful to improve their work. Before we started the annotation each participant signed a consent form and in the instructions it was explicitly mentioned how their data will be used in the work. The data collection protocol is approved by an ethics review board.
Annotators were recruited via using email, resulting in a total of 10 annotators, including 4 females and 6 males, all of them residents of the USA. Annotators were compensated by sharing the results of the study, which proved beneficial in enhancing their work. The instructions of the task include disclaimers of toxic comments which are provided to the annotators. Prior to annotation, each participant provided informed consent by signing a consent form. Participants were aware of how the data would be used. The data collection protocol underwent ethical review board approval.
\subsection{Toxicity and Toxicity Sub-type Definition}
 \begin{itemize}
\item Toxicity is defined as ``a rude, disrespectful, or unreasonable comment that is likely to make people leave a discussion''. This is measured on a 4-point Likert scale with values between -2 and 1, where -2 = Very toxic, -1 = Toxic, 0 = Unsure, and 1 = Not toxic.  
 \item Profanity or Obscenity is defined as ``swear words, curse words, or other obscene or profane language''. This is measured on a 3-point Likert scale from -1 to 1.
\item Insult is defined as ``insulting, inflammatory, or negative comment towards a person or a group of people''. This is measured on a 3-point Likert scale from -1 to 1.
\item Threatening describes ``an intention to inflict pain, injury, or violence against an individual or group''. This is measured on a 3-point Likert scale from -1 to 1.
 \item Identity-based negativity is defined as ``negative or hateful comments targeting someone because of their identity''. This is measured on a 3-point Likert scale from -1 to 1.

 \end{itemize}

 % \subsection{Intended Use of 2023 dataset (}

\subsection{Dataset transformation for calculating of IRR and xRR}
In terms of the 2017 and 2022 datasets, we take into account 25,500 comments which are common in both these datasets. The 2017 dataset has a varying number of annotators for each comment whereas for the 2022 dataset, each group has 5 annotators. Therefore we consider random 5 annotators for each comment from the 2017 dataset. 
 We transform the 2023 dataset (expert annotators) dataset and the 2022 dataset to binary scale since the 2017 dataset is in binary scale in order to maintain consistency and quantitatively analyze all three datasets in an effective manner. 
% \section{Disagreement across 2017 groups nad 2022 subgroups}

% Table \ref{table:XRR1} shows disagreement across all possible combination 2022 subgroups and Table \ref{table:XRR2} shows disagreement across all possible combinations of 2017 and 2022 subgroups.

% \section  {High Disagreement between expert annotators for toxicity Example} 
% Table \ref{table:example_high_disagreement_inside_each_team} shows examples of high disagreement inside each expert team. 

% The variance of each comment in Table \ref{table:example_high_disagreement_inside_each_team} is in the range of threshold value as mentioned above.

% \section{ High disagreement examples across two pools of experts Example}
% Table \ref{table:example_high_disagreement_among_teams}, where the annotations of two expert pools are shown along with scaled differences within the threshold value as defined above.

% \section{ Qualitative analysis example}

% Table \ref{table:disagreement} shows examples which represent five themes for annotators' disagreement along with their scaled difference.

\section{Simulated Toxic Dataset Experiment Details}
\label{sec:appendix_simulated}

\subsection{Simulated dataset}
To simulate a dataset, we first create an entire matrix of ratings from $N$ annotators and $M$ items, resulting in a fully rated dataset. We then reduce the replication to more realistic levels by randomly sampling down ratings per item. If the replication is $P$ that refers to $M$ items annotated by $P$ annotators. 
Then we establish a set of attributes(e.g.\ annotator skill level and comment difficulty level). For annotators' skill level, there are 4 attributes : (1) Expert (2) Average (3) Bad (4) Random whereas in the case of comment difficulty level, there are 3 attributes : (1) Easy (2) Normal (3) Hard. They are described as follows: 
 % \subsection{Annotator Skill Level and Comment Difficulty level}
% begin{itemize}
    \item Annotator skill level:
     \begin{itemize}
         \item Expert: Annotators whose annotation is 90\% accurate with respect to the ground truth
         \item Average: Annotators whose annotation is 75\% accurate with respect to the ground truth
         \item Bad: Annotators whose annotation is 25\% accurate with respect to ground truth
         \item Random: Annotators who annotate 50\% of the comments as toxic or not toxic uniformly at random
         % \item Lazy - represent the type of raters who annotate 90\% of the comments toxic or not toxic 
         % \item Sensitive/Insensitive - represent the type of raters who annotate 25\% of the comments toxic or not toxic
        % \item Biased - Perform differently on items. Can be combined with the skill types above
     \end{itemize}
\end{itemize}

\begin{itemize}
    \item Comment difficulty level:
        \begin{itemize}
            \item Easy: Annotators perform at 120\% of their skill 
            \item Normal: Annotators perform as expected 
            \item Hard: Annotators perform at 70\% of their skill
        \end{itemize}
\end{itemize}

We randomly initialize attributes to each annotator and item and use these attributes to probabilistically assign an annotation for each annotator and item interaction. We then create items and annotator embeddings using the following way 
We first created a sparse binary matrix of item-annotator items. Then we used WALS factorization \cite{hu2008collaborative} to do the dimensionality reduction to create items and annotator embeddings. 
Then we use HBDSCAN for clustering that.

\subsection{Ratio Details}
(1) Ground truth ratio is 1:1 for toxic: not toxic
(2) Annotator type ratio is 1:1:1:5 for expert:random:bad:average.
(3) Item type ratio is 1:2:1 for easy:normal:hard.

% We randomly initialize attributes to each annotator and item. We then use these attributes to probabilistically assign an annotation for each annotator and item interaction. We use the following clustering approach to create items and annotator embeddings:
% We use a standard annotator-item co-clustering approach using the (sparse) ratings as input. 

\subsection{Item-Annotator Embedding}
We first created a sparse binary matrix of item-annotator items. Then we used WALS factorization \cite{hu2008collaborative} to do the dimensionality reduction to create items and annotator embeddings. 
% Then we use HBDSCAN for clustering that.
This formalizes the idea that two annotators are likely to be more similar if they both annotate the same item similarly. And analogously, two items are likely to be more similar if an annotator annotates the same item similarly.
 \subsection{Annotators and Item embedding Error in experiment}
 Positive Train: 0.71, Negative Train: 0.72, Positive Dev: 0.75, Negative Dev: 0.76.

\subsection{Different options in the experiment to improve results %make improvements 
with lower replication}
\begin{itemize}
    \item Scale/normalize/whiten the embedding before running UMAP: Scaling and normalizing correspond to subtracting the mean and normalizing to unit variance, while whitening means removing correlations among features.
    \item Reduced the dev/train error ratio: In the original run, we used 3 dimensions, reg 0.1, and 5 iterations. We tried to perturb dimensions, reg, and iterations to see the scope of improvement. We looked in the range dim =[2,3,5], reg = [0.01, 0.1,1, 10] and iterations =[2,5,8] and saw that dimension 3, reg=10 and iteration = 5 work better which results in the following train/dev error: Positive Train: 0.68, Negative Train: 0.69, Positive Dev: 0.80, and Negative Dev: 0.81.
    \item Trying different UMAP parameters: We modified the UMAP parameter of neighbors, which was set at 15 by default. We selected 30 because lower numbers result in a more dense chain-like structure, whereas higher values (e.g., neighbors = 30) result in a less dense chain-like structure, which helps the clustering method to perform better.

\end{itemize}
\subsection{Effect of majority vote}
\begin{figure}
    \centering
    \includegraphics[width=0.48\textwidth]
    {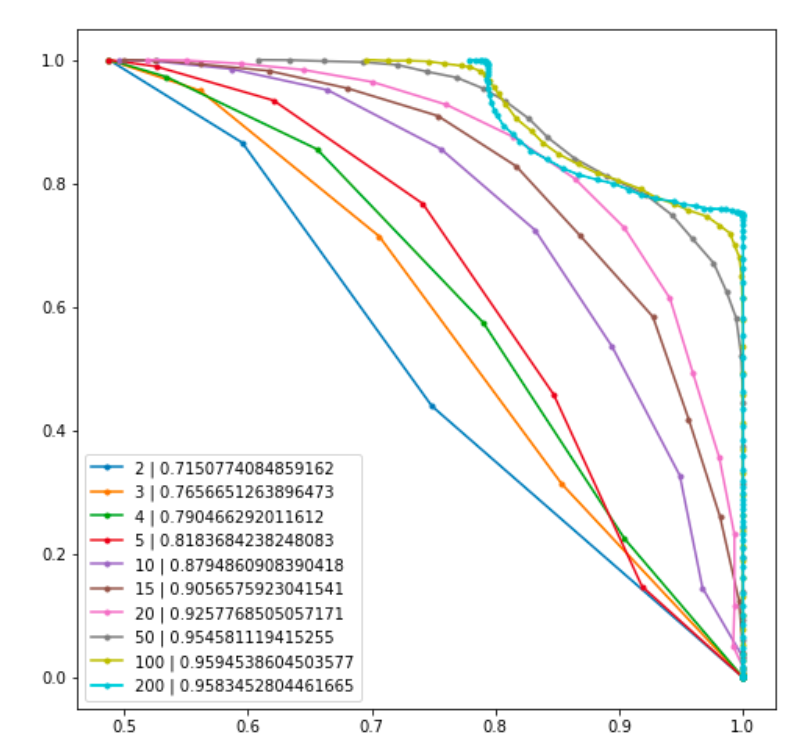}
\caption{Effects of majority vote as a proxy ground truth by replication size}
\label{fig:different_replication_effect}
\end{figure}

Figure~\ref{fig:different_replication_effect} shows When a majority vote is used as ground truth then the AUC decreases as the replication decreases.

\section{LLM Experiment Details}
\subsubsection{Model Training}
\label{sec:appendix_model_training}

We train our LLM using soft prompt tuning \cite{lester-etal-2021-power}. This allows us to efficiently tune the LLM on 10 - 500 examples. The LLM we use is the 62b FLAN-cont-PaLM \cite{flan_palm}. We tuned a prompt consisting of 100 tokens, each with an embedding of dimension 8192 for the 62B model. We trained these tokens with a basic Adam optimizer with clipped gradients. We chose a default learning rate and schedule for all our experiments and did not tune hyperparameters further. We set the sampling temperature to 0 and fixed the prompt initialization, to reduce randomness. For each training run, we have a train, validation, and test set with balanced positive and negative classes. We train for 20-40 epochs and use a validation set to pick the final model for each run. We follow the setup described in \cite{mittal2023using} for the construction of the few shots and soft tuning. We used sklearn for accuracy metrics, whereas TensorFlow internals were used for the model training and implementation.

\end{document}